\pdfoutput=1
\documentclass[11pt,a4paper]{article}
\usepackage[hyperref]{acl2021}
\usepackage{times}
\usepackage{latexsym}

\usepackage{multirow}
\usepackage{multibib}
\usepackage{booktabs}

\usepackage{color}
\usepackage{tikz}
\usepackage{pgfplots}
\usepackage{graphicx}
\usepackage{subfig}
\usepackage{caption}
\usepackage{amsmath}
\usepackage{makecell}
\usepackage{hyperref}

\usetikzlibrary{plotmarks}
\usetikzlibrary{arrows}
\usetikzlibrary{decorations}
\usetikzlibrary{decorations.pathreplacing}
\usetikzlibrary{decorations.markings}
\usetikzlibrary{backgrounds}
\usetikzlibrary{fit}
\usetikzlibrary{positioning}
\usetikzlibrary{calc}
\usetikzlibrary{patterns}
\usetikzlibrary{shadows}
\usetikzlibrary[shapes.multipart]
\usetikzlibrary{pgfplots.groupplots}

\definecolor{ugreen}{cmyk}{1,0,1,0.498}
\definecolor{dblue}{cmyk}{1,0.5487,0,0.5569}

\definecolor{ugreen}{rgb}{0,0.5,0}
\definecolor{mygreen}{RGB}{46,139,87}
\definecolor{iyellow}{RGB}{255,250,205}
\definecolor{ipurple}{RGB}{230,230,250}
\definecolor{myred}{RGB}{238,44,44}
\definecolor{myblue}{RGB}{30,144,255}
\definecolor{myorange}{RGB}{255,127,80}
\definecolor{mypurple}{RGB}{255,20,147}

\usepackage{microtype}

\aclfinalcopy 

\title{Stacked Acoustic-and-Textual Encoding: Integrating the Pre-trained Models into Speech Translation Encoders}

\author{
  Chen Xu\textsuperscript{1},
  Bojie Hu\textsuperscript{2},
  Yanyang Li\textsuperscript{3},
  Yuhao Zhang\textsuperscript{1}, \\
  \textbf{Shen Huang\textsuperscript{2},
  {Qi Ju\textsuperscript{2}}\thanks{\ \ Corresponding author} \ ,
  Tong Xiao\textsuperscript{1,4$\ast$},
  Jingbo Zhu\textsuperscript{1,4}} \\
  \textsuperscript{1}NLP Lab, School of Computer Science and Engineering, \\
    Northeastern University, Shenyang, China \\
  \textsuperscript{2}Tencent Minority-Mandarin Translation, Beijing, China \\
  \textsuperscript{3}The Chinese University of Hong Kong, Hong Kong, China \\
  \textsuperscript{4}NiuTrans Research, Shenyang, China \\
  {\tt \{xuchenneu,blamedrlee,yoohaozhang\}@outlook.com}, \\
  {\tt \{bojiehu,springhuang,damonju\}@tencent.com}, \\
  {\tt \{xiaotong,zhujingbo\}@mail.neu.edu.cn} \\
}

\date{}

\begin{document}
\maketitle
\begin{abstract}

Encoder pre-training is promising in end-to-end Speech Translation (ST), given the fact that speech-to-translation data is scarce.
But ST encoders are not simple instances of Automatic Speech Recognition (ASR) or Machine Translation (MT) encoders.
For example, we find that ASR encoders lack the global context representation, which is necessary for translation, whereas MT encoders are not designed to deal with long but locally attentive acoustic sequences.
In this work, we propose a \emph{Stacked Acoustic-and-Textual Encoding} (SATE) method for speech translation.
Our encoder begins with processing the acoustic sequence as usual, but later behaves more like an MT encoder for a global representation of the input sequence.
In this way, it is straightforward to incorporate the pre-trained models into the system.
Also, we develop an adaptor module to alleviate the representation inconsistency between the pre-trained ASR encoder and MT encoder, and develop a multi-teacher knowledge distillation method to preserve the pre-training knowledge.
Experimental results on the LibriSpeech En-Fr and MuST-C En-De ST tasks show that our method achieves state-of-the-art BLEU scores of 18.3 and 25.2.
To our knowledge, we are the first to develop an end-to-end ST system that achieves comparable or even better BLEU performance than the cascaded ST counterpart when large-scale ASR and MT data is available\footnote{The source code is available at \href{https://github.com/xuchenneu/SATE}{https://github.com/xuchen\\neu/SATE}}.

\end{abstract}

\section{Introduction}

End-to-end Speech Translation (E2E ST) has become popular recently for its ability to free designers from cascading different systems and shorten the pipeline of translation \cite{Duong_naacl2016,Berard_arxiv2016,Weiss_ISCA2017}.
Promising results on small-scale tasks are generally favorable. However, speech-to-translation paired data is scarce.
Researchers typically use pre-trained Automatic Speech Recognition (ASR) and Machine Translation (MT) models to boost ST systems \cite{Berard_IEEE2018}. For example, one can initialize the ST encoder using a large-scale ASR model \cite{Bansal_NAACL2019}. But we note that, despite significant development effort, our end-to-end ST system with pre-trained models was not able to outperform the cascaded ST counterpart when the ASR and MT data size was orders of magnitude larger than that of ST (see Table \ref{table_intro}).

\begin{table}[t]
  \centering
  \begin{tabular}{l|c|c}
    \toprule
    Setting & Model & BLEU \\
    \midrule
    \multirow{2}{*}{Restricted} & Cascaded & 23.3 \\
     & E2E+Pre-training & 23.1 \\
    \midrule
    \multirow{2}{*}{Unrestricted} & Cascaded & 28.1 \\
    & E2E+Pre-training & 25.6 \\
    \bottomrule
  \end{tabular}
  \caption{BLEU scores [\%] of a cascaded ST model and an end-to-end ST model with pre-training on the MuST-C En-De corpus. Restricted = training is restricted to the ST data, and Unrestricted = additional training data is allowed for ASR and MT.}
  \label{table_intro}
\end{table}

In this paper, we explore reasons why pre-training has been challenging in ST, and how pre-trained ASR and MT models might be used together to improve ST.
We find that the ST encoder plays both roles of acoustic encoding and textual encoding. This makes it problematic to view an ST encoder as either an individual ASR encoder or an individual MT encoder. More specifically, there are two problems.

\begin{itemize}
  \item Modeling deficiency: the MT encoder tries to capture long-distance dependency structures of language, but the ASR encoder focuses more on local dependencies in the input sequence. Since the ST encoder is initialized by the pre-trained ASR encoder \cite{Berard_IEEE2018}, it fails to model large contexts in the utterance. But a large scope of representation learning is necessary for translation \cite{Yang_emnlp2018}.
  \item Representation inconsistency: on the decoder side of ST, the MT decoder is in general used to initialize the model. The assumption here is that the upstream component is an MT-like encoder, whereas the ST encoder actually behaves more like an ASR encoder.
\end{itemize}

We address these problems by marrying the world of ASR encoding with the world of MT encoding.
We propose a \emph{Stacked Acoustic-and-Textual Encoding} (SATE) method to cascade the ASR encoder and the MT encoder.
It first reads and processes the sequence of acoustic features as a usual ASR encoder.
Then an adaptor module passes the acoustic encoding output to an MT encoder with two principles: informative and adaptive.
In this way, pre-trained ASR and MT encoders can work for what we would originally design them, and the incorporation of pre-trained models into ST is more straightforward.
In addition, we develop a multi-teacher knowledge distillation method to robustly train the ST encoder and preserve the pre-trained knowledge during fine-tuning \cite{Yang_AAAI2020}.

We test our method in a Transformer-based end-to-end ST system.
Experimental results on the LibriSpeech En-Fr and MuST-C En-De speech translation benchmarks show that it achieves the state-of-the-art performance of 18.3 and 25.2 BLEU points.
Under a more challenging setup, where the large-scale ASR and MT data is available, SATE achieves comparable or even better performance than the cascaded ST counterpart.
We believe that we are the first to present an end-to-end system that can beat the strong cascaded system in unrestricted speech translation tasks.

\section{Related Work}

Speech translation aims at learning models that can predict, given some speech in the source language, the translation into the target language.
The earliest of these models were cascaded: they treated ST as a pipeline of running an ASR system and an MT system sequentially \cite{Ney_IEEE1999,Mathias_ICASSP2006,schultz_2004}.
This allows the use of off-the-shelf models, and was (and is) popular in practical ST systems.
However, these systems were sensitive to the errors introduced by different component systems and the high latency of the long pipeline.

As another stream in the ST area, end-to-end methods have been promising recently \cite{Berard_arxiv2016, Weiss_ISCA2017,Berard_IEEE2018}.
The rise of end-to-end ST can be traced back to the success of deep neural models \cite{Duong_naacl2016}.
But, unlike other well-defined tasks in deep learning, annotated speech-to-translation data is scarce, which prevents well-trained ST models.
A simple solution to this issue is data augmentation \cite{pino_corr2019,Pino_ISCA2020}.
This method is model-free but generating large-scale synthetic data is time consuming.
As an alternative, researchers used multi-task learning (MTL) to robustly train the ST model so that it could benefit from additional guide signals \cite{Weiss_ISCA2017,Anastasopoulos_NAACL2018,Berard_IEEE2018, Sperber_tacl2019, Dong_aaai2021}.
Generally, MTL requires a careful design of the loss functions and more complicated architectures.

In a similar way, more recent work pre-trains different components of the ST system, and consolidates them into one.
For example, one can initialize the encoder with an ASR model, and initialize the decoder with the target-language side of an MT model \cite{Berard_IEEE2018,Bansal_NAACL2019,Stoian_ICASSP2020}.
More sophisticated methods include better training and fine-tuning \cite{Wang_aaai2020,Wang_acl2020}, the shrink mechanism \cite{Liu_corr2020}, the adversarial regularizer \cite{Alinejad_EMNLP2020}, and etc.
Although pre-trained models have quickly become dominant in many NLP tasks, they are still found to underperform the cascaded model in ST.
This motivates us to explore the reasons why this happens and methods to solve the problems accordingly.

\begin{figure*}[t!]
  \begin{tikzpicture}
            \footnotesize{
            \begin{axis}[
              ymajorgrids,
              xmajorgrids,
              grid style=dashed,
              width=.30\textwidth,
              height=.30\textwidth,
              legend columns=3,
              legend entries={ST, ASR, MT},
              legend style={
                draw=none,
                line width=1pt,
              },
              legend style={at={(0.5,1.0)}, anchor=south},
              xmin=0, xmax=13,
              ymin=0, ymax=1,
              xtick={0, 4, 8, 12},
              ytick={0.2, 0.4, 0.6, 0.8},
              xlabel=\footnotesize{Layer},
              ylabel=\footnotesize{Localness},
              ylabel style={yshift=-1em},
              xlabel style={yshift=0.0em},
              yticklabel style={/pgf/number format/precision=2,/pgf/number format/fixed zerofill},
              scaled ticks=false,
              ]
          \addplot[mygreen!80, mark=triangle, line width=1pt] file {data/st.txt};
          \addplot[myblue!80, mark=star, line width=1pt] file {data/asr.txt};
          \addplot[myred!80, mark=diamond*, line width=1pt] file {data/mt.txt};
          \end{axis}
          }
  \node [anchor=center](pos1)at (1.6,-1.4){\footnotesize{(a) Localness Analysis}};
  \footnotesize{
          \begin{axis}[
  xshift=16.0em,
            ybar=1pt,
            bar width=4pt,
            ymajorgrids,
            xmajorgrids,
            grid style=dashed,
            width=.3\textwidth,
            height=.30\textwidth,
            xlabel=\footnotesize{CTC Position},
            ylabel=\footnotesize{Localness},
            ylabel style={yshift=-1em},
            xlabel style={yshift=0.0em},
            yticklabel style={/pgf/number format/precision=2,/pgf/number format/fixed zerofill},
            legend columns=3,
            legend style={
              draw=none,
              line width=1pt,
            },
            xmin=0, xmax=13,
            ymin=0.40, ymax=0.65,
            xtick={0, 4, 8, 12},
            ytick={0.45, 0.50, 0.55, 0.60},
            tick align=inside,
            scaled ticks=false,
            legend style={at={(0.5,1.0)}, anchor=south},
            legend image code/.code={
                \draw [#1] (0cm,-0.1cm) rectangle (0.3cm,0.1cm);
            },
            ]
        \addplot[color=myblue,fill=myblue!60!white, line width=1pt] coordinates {
        (2, 0.63) (4, 0.62) (6, 0.59) (8, 0.56) (10, 0.55)
        };
        \addlegendentry{Below CTC} 
        \addplot[color=myred,fill=myred!60!white, line width=1pt] coordinates {
        (2, 0.47) (4, 0.46) (6, 0.46) (8, 0.49) (10, 0.48)
        };
        \addlegendentry{Above CTC} 
        \end{axis}
        }
  \node [anchor=west](pos2)at ([xshift=4.7em]pos1.east){\footnotesize{(b) CTC Impact on Localness}};
  \footnotesize{
          \begin{axis}[
  xshift=32.0em,
            ymajorgrids,
            xmajorgrids,
            grid style=dashed,
            width=.30\textwidth,
            height=.30\textwidth,
            xlabel=\footnotesize{CTC Position},
            ylabel=\footnotesize{BLEU(ST)},
            ylabel style={yshift=-1em},
            xlabel style={yshift=0.0em},
            yticklabel style={/pgf/number format/precision=1,/pgf/number format/fixed zerofill},
            legend columns=3,
            legend style={
              draw=none,
              line width=1pt,
            },
            xmin=0, xmax=13,
            ymin=19.8, ymax=22,
            xtick={0, 4, 8, 12},
            ytick={20.0, 20.5, 21.0, 21.5, 22.0},
            scaled ticks=false,
            legend style={at={(0.5,1)}, anchor=south},
            ]
        \addplot[mark=triangle, color=mygreen!80, line width=1pt] coordinates {
          (2, 20.05) (4, 21.14) (6, 21.59) (8, 21.77) (10, 21.49) (12, 21.46)
          };\label{BLEU}
        \end{axis}
        \begin{axis}[
  xshift=32.0em,
          ymajorgrids,
          xmajorgrids,
          grid style=dashed,
          width=.30\textwidth,
          height=.30\textwidth,
          ylabel=\footnotesize{WER(ASR)},
          ylabel style={yshift=-1em},
          xlabel style={yshift=0.0em},
          yticklabel style={/pgf/number format/precision=1,/pgf/number format/fixed zerofill},
          xtick=\empty,
          legend columns=3,
          legend style={
            draw=none,
            line width=1pt,
          },
          axis lines=right,
          axis line style={-},
          tick align=inside,
          xmin=0, xmax=13,
          ymin=11.3, ymax=13.5,
          ytick={11.5, 12.0, 12.5, 13.0, 13.5},
          scaled ticks=false,
          legend style={at={(0.5,1)}, anchor=south},
        ]
        \addlegendimage{/pgfplots/refstyle=BLEU}\addlegendentry{ST}
        \addplot[mark=star, color=myblue!80, line width=1pt] coordinates {
          (2, 13.3) (4, 12.8) (6, 12) (8, 12.1) (10, 12.6) (12, 11.8)
          };\addlegendentry{ASR}
        \end{axis}
          }
  \node [anchor=west](pos3)at ([xshift=3.2em]pos2.east){\footnotesize{(c) CTC Impact on Performance}};
      \end{tikzpicture}
    \caption{(a) Localness in each layer of the ST, ASR, and MT encoders, (b) the impact of CTC position on localness, and (c) the impact of CTC position on performance of ST and ASR models.}
    \label{fig:analysis}
  \end{figure*}
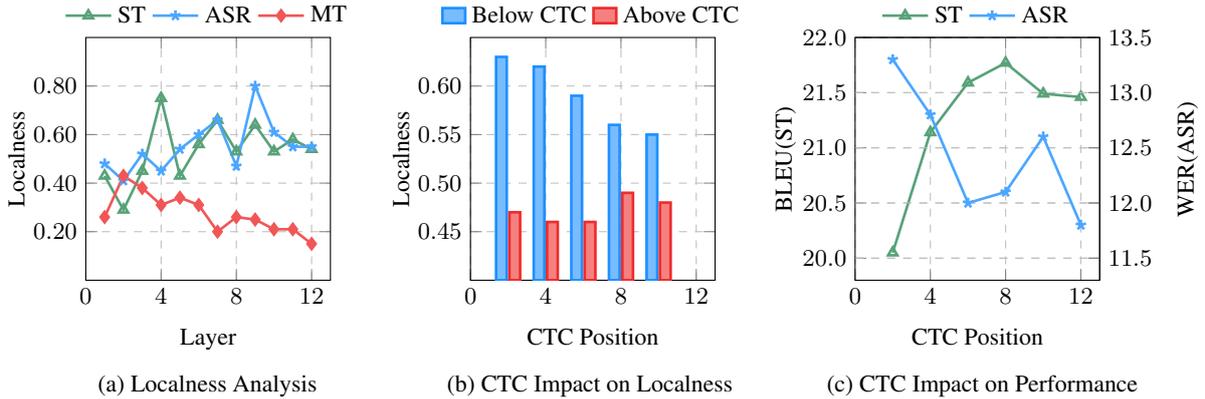

\section{Why is ST Encoding Difficult?}
\label{section_3}

Following previous work in end-to-end models \cite{Berard_arxiv2016,Weiss_ISCA2017}, we envision an encoding-decoding process in which an input sequence is encoded into a representation vector, and the vector is then decoded into an output sequence.
In such a scenario, all end-to-end ST, ASR and MT systems can be viewed as instances of the same architecture.
Then, components of these systems can be pre-trained and re-used across them.

An underlying assumption here is that the ST encoder is doing something quite similar to what the MT (or ASR) encoder is doing.
However, \newcite{Sperber_ISCA2018} find that the ASR model benefits from a small attention window, which is inconsistent with the MT model \cite{Yang_emnlp2018}.
To verify this, we compare the behavior of ST, ASR and MT encoders.
We choose Transformer as the base architecture \cite{Vaswani_nips2017} and run experiments on the MuST-C En-De corpus.
We report the results on the MuST-C En-De tst-COMMON test data.
For stronger systems, we use Connectionist Temporal Classification (CTC) \cite{Graves_ACL2006} as the auxiliary loss on the encoders when we train the ASR and ST systems \cite{Watanabe_IEEE2017,Karita_ISCA2019,Bahar_ASRU2019}.
The CTC loss forces the encoders to learn alignments between speech and transcription.
It is necessary for the state-of-the-art performance \cite{Watanabe_ISCA2018}.

Here we define the \emph{localness} of a word as the sum of the attention weights to the surrounding words (or features) within a fixed small window\footnote{Here we treat the attention weight of Transformer as a distribution over all positions.}.
The window size is $10\%$ of the sequence length.
Figure \ref{fig:analysis}(a) shows the localness of the attention weights for different layers of the encoders. We see that the ST and ASR encoders prefer local attention which indicates a kind of short-distance dependencies in processing acoustics feature sequences. Whereas the MT encoder generates a more global distribution of attention weights for word sequences, especially when we stack more layers. This result arises a new question: Is local attention sufficient for speech translation?

Then, we design another experiment to examine if the high localness in attention weights of the ASR and ST encoders is due to the bias imposed by CTC.
In Figure \ref{fig:analysis}(b), we use the CTC loss in the intermediate layer and show the average localness of the layers above or below CTC.
The CTC loss demonstrates strong preference for locally attentive models.
The upper-level layers act more like an MT encoder, that is, the layers with no CTC loss generates more global distributions.
Taking this further, Figure \ref{fig:analysis}(c) demonstrates a slightly higher BLEU score when we free more upper-level layers from the guide of CTC.
Meanwhile, the word error rate (WER) increases because only lower parts of the model are learned in a standard manner of ASR.

Now we have some hints: the ST encoder is not a simple substitution of the ASR encoder or the MT encoder. Rather, they are complementary to each other, that is, we need the ASR encoder to deal with the acoustic input, and the MT encoder to generate the representation vector that can work better with the decoder.

\section{The Method}

In speech translation, we want the encoder to represent the input speech to some sort of decoder-friendly representations. We also want the encoder to be ``natural'' for pre-training. In the following, we describe, \emph{Stacked Acoustics-and-Textual Encoding} (SATE), a new ST encoding method to meet these requirements, and improvements of it.

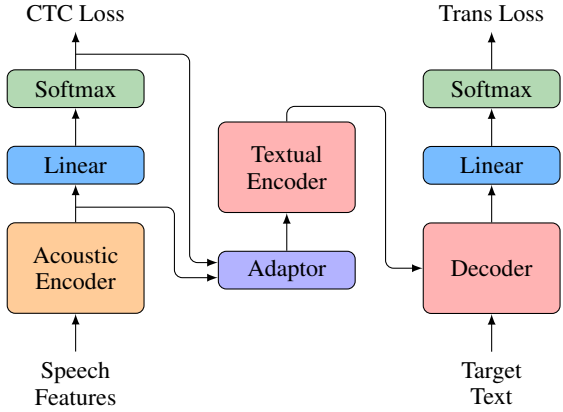
\begin{figure}
  \begin{center}
    \begin{tikzpicture}
      \tikzstyle{model} = [rectangle,draw,minimum width=1.8cm,font=\small,rounded corners=3pt,align=center,inner sep=1pt]
      \tikzstyle{textonly} = [font=\small,align=center]

      \node[textonly] (speech) at (0,0) {Speech\\Features};
      \node[model,minimum height=1.2cm,anchor=south,fill=orange!40] (asr) at ([yshift=0.5cm]speech.north) {Acoustic\\Encoder};
      \node[model,minimum height=0.5cm,anchor=south,fill=myblue!60] (linear1) at ([yshift=0.5cm]asr.north) {Linear};
      \node[model,minimum height=0.5cm,anchor=south,fill=ugreen!30] (softmax1) at ([yshift=0.5cm]linear1.north) {Softmax};
      \node[textonly,anchor=south] (ctc) at ([yshift=0.5cm]softmax1.north) {CTC Loss};

      \draw[-latex] (speech) -- (asr);
      \draw[-latex] (asr) -- (linear1);
      \draw[-latex] (linear1) -- (softmax1);
      \draw[-latex] (softmax1) -- (ctc);

      \node[model,minimum height=0.5cm,anchor=west,fill=blue!30] (bridge) at ([xshift=1.2cm,yshift=1.5cm]speech.east) {Adaptor};
      \node[model,minimum height=1.2cm,anchor=south,fill=red!30] (enc) at ([yshift=0.5cm]bridge.north) {Textual\\Encoder};

      \draw[-latex] (bridge) -- (enc);

      \draw[-latex,rounded corners=3pt] ([yshift=0.2cm]softmax1.north) -- +(1.5cm,0) |- ([yshift=0.1cm]bridge.west);
      \draw[-latex,rounded corners=3pt] ([yshift=0.2cm]asr.north) -- +(1.3cm,0) |- ([yshift=-0.1cm]bridge.west);

      \node[textonly,anchor=west] (target) at ([xshift=2.4cm+1.8cm+2pt]speech.east) {Target\\Text};
      \node[model,minimum height=1.2cm,anchor=south,fill=red!30] (dec) at ([yshift=0.5cm]target.north) {Decoder};
      \node[model,minimum height=0.5cm,anchor=south,fill=myblue!60] (linear2) at ([yshift=0.5cm]dec.north) {Linear};
      \node[model,minimum height=0.5cm,anchor=south,fill=ugreen!30] (softmax2) at ([yshift=0.5cm]linear2.north) {Softmax};
      \node[textonly,anchor=south] (mt) at ([yshift=0.5cm]softmax2.north) {Trans Loss};

      \draw[-latex] (target) -- (dec);
      \draw[-latex] (dec) -- (linear2);
      \draw[-latex] (linear2) -- (softmax2);
      \draw[-latex] (softmax2) -- (mt);

      \draw[-latex,rounded corners=3pt] (enc.north) -- ([yshift=0.2cm]enc.north) -- +(1.3cm,0) |- (dec.west);

    \end{tikzpicture}
  \end{center}
  \caption{The overall architecture of stacked acoustic-and-textual encoding.}
  \label{arch}
\end{figure}

\subsection{Stacked Acoustic-and-Textual Encoding}
Unlike previous work, the SATE method does not rely on a single encoder to receive the signal from both the CTC loss and the feedback of the decoder. Instead, it is composed of two encoders: the first does exactly the same thing as the ASR encoder (call it \textit{acoustic encoder}), and the other generates a higher-level globally-attentive representation on top of the acoustic encoder (call it \textit{textual encoder}).

See Figure \ref{arch} for the architecture of SATE. The acoustic encoder is trained by CTC in addition to the supervision signal from the translation loss. Let $(x, y^s, y^t)$ be an ST training sample, where $x$ is the input feature sequence of the speech, $y^s$ is the transcription of $x$, and $y^t$ is the translation in the target language. We define the output of the acoustic encoder as:
\begin{eqnarray}
h^s & = & E_{s}(x)
\end{eqnarray}

\noindent where $E_{s}(\cdot)$ is the encoding function. Then, we add a Softmax layer on $h^s$ to predict the CTC label path $\pi = (\pi_1, \cdots, \pi_{T})$, where $T$ is the length of the input sequence. The probability of path $\textrm{P}(\pi | h^s)$ is the product of the probability $\textrm{P}(\pi_t | h^s_t)$ at every time $t$ based on conditionally independent assumption:
\begin{eqnarray}
  \textrm{P}(\pi | h^s) & \approx & \prod_{t}^{T} \textrm{P}(\pi_{t} | h^s_t)
  \label{ctc_prob}
\end{eqnarray}

CTC works by summing over the probability of all possible alignment paths $\Phi(y^s)$ between $x$ and $y^s$ , as follows:
\begin{eqnarray}
\textrm{P}_{\rm CTC}(y^s|h^s) & = & \sum_{\pi \in \Phi(y^s)} \textrm{P}(\pi | h^s)
\end{eqnarray}

Then, the CTC loss is defined as:
\begin{eqnarray}
\mathcal{L}_{\rm CTC} & = & -\log \textrm{P}_{\rm CTC}(y^s | h^s; \theta_{\rm CTC})
\end{eqnarray}

\noindent where $\theta_{\rm CTC}$ is the model parameters of the acoustic encoder and the CTC output layer.

The acoustic encoder is followed by an adaptor. It receives $h^s$ and $P(\pi| h^s)$, and produces  a new representation required by the textual encoder. Let $A(\cdot,\cdot)$ be the adaptor module. Its output is defined as:
\begin{eqnarray}
\hat{h^s} & = & A(h^s, \textrm{P}(\pi | h^s))
\end{eqnarray}

\noindent We leave the design of the adaptor to Section \ref{sec:adaptor}.
Furthermore, we stack the textual encoder on the adaptor. The output $h^t$ is defined as:
\begin{eqnarray}
h^t       & = & E_t(\hat{h^s})
\end{eqnarray}

\noindent where $E_{t}(\cdot)$ is the textual encoder. $h^t$ is fed into the decoder for computing the translation probability $\textrm{P}_{\rm Trans} (y^t | h^t)$, as in standard MT systems. We define the translation loss as:
\begin{eqnarray}
\mathcal{L}_{\rm Trans} &= -\log \textrm{P}_{\rm Trans} (y^t | h^t; \theta_{\rm ST})
\end{eqnarray}

\noindent where $\theta_{\rm ST}$ is all model parameters except for the CTC output layer.

Finally, we interpolate $\mathcal{L}_{\rm CTC}$ and $\mathcal{L}_{\rm Trans}$ (with coefficient $\alpha$) for the loss of the entire model:
\begin{eqnarray}
\mathcal{L} & = & \alpha \cdot \mathcal{L}_{\rm CTC} + (1 - \alpha) \cdot \mathcal{L}_{\rm Trans} \label{eq:loss}
\end{eqnarray}

Since the textual encoder works for the decoder only, it is trained as an MT encoder. In this way, the acoustic and textual encoders can do what we would originally expect them to do: the acoustic encoder deals with the acoustic input (i.e., ASR encoding), and the textual encoder generates a representation for translation (i.e., MT encoding).
Also, SATE is friendly to pre-training. One can simply use an ASR encoder as the acoustic encoder, and use an MT encoder as the textual encoder. Note that SATE is in general a cascaded model, in response to the pioneering work in ST \cite{Ney_IEEE1999}. It can be seen as cascading the ASR and MT systems in an end-to-end fashion.
\subsection{The Adaptor}
\label{sec:adaptor}
Now we turn to the design of the adaptor.
Note that the pre-trained MT encoder assumes that the input is a word embedding sequence.
Simply stacking the MT encoder and the ASR encoder obviously does not work well.
For this reason, the adaptor fits the output of the ASR encoder (i.e., the acoustic encoder) to what an MT encoder would like to see.
We follow two principles in designing the adaptor: \textit{adaptive} and \textit{informative}.

We need an adaptive representation to make the input of the textual encoder similar to that of the MT encoder.
To this end, we generate the soft contextual representation that shares the same latent space with the embedding layer of the MT encoder.

As shown in Eq. (\ref{ctc_prob}), the CTC output $\textrm{P} (\pi_t|h^s_t)$ indicates the alignment probability over the vocabulary at time $t$.
Instead of replacing the representation by the embedding of the most-likely token \cite{Liu_corr2020}, we employ a soft token which is the expectation of the embedding over the distribution from CTC.
Let $W^e$ be the embedding matrix of the textual encoder, we define the soft representation $h^s_{\rm soft}$ as:
\begin{eqnarray}
h^s_{\rm soft} & = & \textrm{P}(\pi | h^s) \cdot W^e
\label{eq:soft-h}
\end{eqnarray}



Also, an informative representation should contain information in the original input \cite{Peters_NAACL2018}.
The output acoustic representation of the ASR encoder generally involves paralinguistic information, such as emotion, accent, and emphasis.
They are not expressed in the form of text explicitly but might be helpful for translation.
For example, the generation of the declarative or exclamatory sentences depends on the emotions of the speakers.

We introduce a single-layer neural network to learn to map the acoustic representation to the latent space of the textual encoder, which preserves the acoustic information:
\begin{eqnarray}
h^s_{\rm map} & = & \textrm{ReLU}(W^{\rm map} \cdot h^s + b^{\rm map})
\end{eqnarray}

\noindent where $W^{\rm map}$ and $b^{\rm map}$ are the trainable parameters.

\begin{figure}
  \begin{center}
    \begin{tikzpicture}
      \tikzstyle{prob} = [rectangle,fill=ugreen!40,minimum width=0.3cm,anchor=south west,inner sep=0pt]
      \tikzstyle{embed} = [rectangle,draw,fill=myblue!60,inner sep=0pt,rounded corners=2pt,minimum size=0.3cm,anchor=south west]
      \tikzstyle{textonly} = [font=\footnotesize,align=center,rotate=45,anchor=south west]

      \coordinate (start) at (0,0);

      \node[rectangle,draw,minimum width=2.5cm,rounded corners=3pt,fill=white,font=\small,align=center,anchor=west] (mapping) at ([xshift=1cm]start.east) {Mapping Layer};

      \node[prob,minimum height=0.1cm] (p1) at ([yshift=0.5cm]mapping.north west) {};
      \node[prob,minimum height=0.6cm] (p2) at ([xshift=1pt]p1.south east) {};
      \node[prob,minimum height=0.3cm] (p3) at ([xshift=1pt]p2.south east) {};
      \node[prob,minimum height=0.05cm] (p4) at ([xshift=1pt]p3.south east) {};
      \node[prob,minimum height=0.2cm] (p5) at ([xshift=1pt]p4.south east) {};
      \node[prob,font=\scriptsize,fill=none,minimum height=0.3cm] (p6) at ([xshift=1pt]p5.south east) {$\cdots$};
      \node[prob,minimum height=0.4cm] (p7) at ([xshift=1pt]p6.south east) {};
      \begin{pgfonlayer}{background}
        \node[inner sep=0pt,fit=(p1) (p2) (p3) (p4) (p5) (p6) (p7)] (prob) {};
      \end{pgfonlayer}

      \node[embed] (e1) at ([yshift=0.5cm]prob.north west) {};
      \node[embed] (e2) at ([xshift=1pt]e1.south east) {};
      \node[embed] (e3) at ([xshift=1pt]e2.south east) {};
      \node[embed] (e4) at ([xshift=1pt]e3.south east) {};
      \node[embed] (e5) at ([xshift=1pt]e4.south east) {};
      \node[embed,font=\scriptsize,draw=none,fill=none,minimum height=0.3cm] (e6) at ([xshift=1pt]e5.south east) {$\cdots$};
      \node[embed] (e7) at ([xshift=1pt]e6.south east) {};

      \node[textonly] (w1) at (e1.north) {we};
      \node[textonly] (w2) at (e2.north) {you};
      \node[textonly] (w3) at (e3.north) {I};
      \node[textonly] (w4) at (e4.north) {hi};
      \node[textonly] (w5) at (e5.north) {that};
      \node[textonly] (w7) at (e7.north) {can};

      \node[] (label1) at ([yshift=0.2cm]prob.north) {$\times$};
      \node[anchor=west] (label2) at ([xshift=1.3cm]label1.east) {$=$};
      \node[embed,fill=red!30,anchor=west] (result) at ([xshift=0.2cm]label2.east) {};
      \node[inner sep=0pt,anchor=south] (residual) at ([yshift=0.5cm]result.north) {$\bigoplus$};
      \node[font=\small,anchor=south] (output) at ([yshift=0.7cm]residual.north) {Output};

      \draw[-latex] (mapping.east) -- +(1.7cm,0) |- (residual.east);
      \draw[-latex] (result) -- (residual);
      \draw[-latex] (residual.north) -- (output.south);

      \node[font=\small,anchor=east,align=center,minimum width=2cm] (input1) at ([xshift=-0.5cm]prob.west) {CTC\\Distribution};
      \node[font=\small,anchor=east,align=center,minimum width=2cm] (input2) at ([xshift=-0.5cm]mapping.west) {Acoustic \\Representation};

      \draw[-latex] (input1.east) -- (prob.west);
      \draw[-latex] (input2.east) -- (mapping.west);

      \begin{pgfonlayer}{background}
        \node[] (tmp1) at ([xshift=0.1cm]residual.east) {};
        \node[] (tmp2) at (mapping.west) {};
        \node[rectangle,draw,rounded corners=3pt,inner sep=5pt,fill=gray!10!white,drop shadow,fit=(mapping) (prob) (residual) (result) (w1) (w2) (w3) (w4) (w5) (w7) (tmp1) (tmp2)] (bridge) {};
      \end{pgfonlayer}

      \node[embed,anchor=south east,label={[font=\small,label distance=0pt,name=label3]0:Embedding}] () at ([shift={(-1.9cm,0.3cm)}]bridge.north west) {};
      \node[embed,fill=red!30,anchor=west,label={[font=\small,label distance=0pt,name=label4]0:Soft Embedding}] () at ([xshift=0.2cm]label3.east) {};
    \end{tikzpicture}
  \end{center}
  \caption{The architecture of the adaptor.}
  \label{fig:adaptor}
\end{figure}
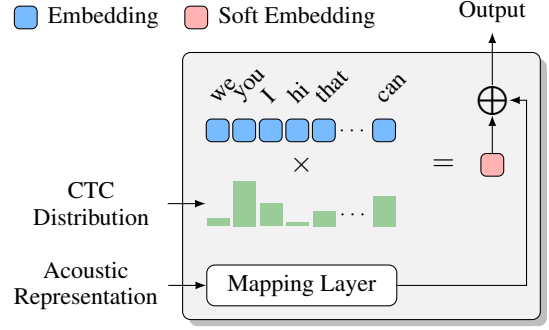

The final output of the adaptor is defined to be:
\begin{eqnarray}
A(h^s, P(\pi| h^s)) &= & \lambda \cdot h^s_{\rm map} + \nonumber \\
                    &  & (1 - \lambda) \cdot h^s_{\rm soft}
\end{eqnarray}
\noindent where $\lambda$ is the weight of $h^s_{\rm map}$ and set to 0.5 by default. Figure \ref{fig:adaptor} shows the architecture of the adaptor.

Note that, in the adaptor, we do not change the sequence length for textual encoding because such a way is simple for implementation and shows satisfactory results in our experiments.
Although there is a length inconsistency issue, the sequence representation of the speech should be similar with the correspond transcription.
Shrinking the sequence simply results in information incompleteness.
We will investigate this issue in the future.

\subsection{Multi-teacher Knowledge Distillation}


Another improvement here is that we develop a multi-teacher knowledge distillation (MTKD) method to preserve the pre-trained knowledge during fine-tuning \cite{Hinton_Corr2015}.


The ST model mimics the teacher distribution by minimizing the cross-entropy loss between the teacher and student \cite{Liu_ISCA2019}.
For a training sample $(x, y^s, y^t)$, we define two loss functions:
\begin{eqnarray}
\mathcal{L}_{\rm KD\_CTC} & = &-\sum_{m=1}^{T}\sum_{k=1}^{|V|} \textrm{Q}(\pi_m=v_k|x; \theta_{\rm ASR}) \nonumber \\
& & \times \log \textrm{P}(\pi_m=v_k|x; \theta_{\rm CTC}) \\
\mathcal{L}_{\rm KD\_Trans} & = &-\sum_{n=1}^{|y^t|}\sum_{k=1}^{|V|} \textrm{Q}(y_n^t=v_k|y^s; \theta_{\rm MT}) \nonumber \\
& & \times \log \textrm{P}(y_n^t=v_k|x; \theta_{\rm ST})
\end{eqnarray}

\noindent where $v_k$ is the word indexed by $k$ and $V$ is the vocabulary shared among the ST, ASR, and MT models.
$\textrm{Q}(\cdot|\cdot)$ is the teacher distribution and $\textrm{P}(\cdot|\cdot)$ is the student distribution.
$\theta_{\rm ASR}$, $\theta_{\rm CTC}$, $\theta_{\rm MT}$ and $\theta_{\rm ST}$ are the model parameters.

We can rewrite Eq. (\ref{eq:loss}) to obtain a new loss:
\begin{eqnarray}
\mathcal{L} & = & \alpha \cdot \big(\beta \cdot  \mathcal{L}_{\rm CTC} + (1 - \beta) \cdot \mathcal{L}_{\rm KD\_CTC} \big) \nonumber \\
            &   & + (1 - \alpha) \cdot   \nonumber \\
            &   & \big(\gamma \cdot \mathcal{L}_{\rm Trans} + (1 - \gamma) \cdot \mathcal{L}_{\rm KD\_Trans} \big) \label{eq:new-loss}
\label{eq:final_obj}
\end{eqnarray}

\noindent where both $\beta$ and $\gamma$ are the hyper-parameters that balance the preference between the teacher distribution and the ground truth.

\section{Experiments}

\subsection{Datasets and Preprocessing}

We consider restricted and unrestricted settings on speech translation tasks. We run experiments on the LibriSpeech English-French (En-Fr) \cite{Kocabiyikoglu_LREC18} and MuST-C English-German (En-De) \cite{DiGangi_NAACL2019} corpora, which correspond to the low-resource and high-resource datasets respectively.
Available ASR and MT data is only from the ST data under the restricted setting.
For comparison in practical scenarios, the unrestricted setting allows the additional data for ASR and MT models.

\noindent \textbf{LibriSpeech En-Fr} Followed previous work, we use the clean speech translation training set of 100 hours, including 45K utterances and doubled translations of \textit{Google Translate}.
We select the model on the dev set (1,071 utterances) and report results on the test set (2,048 utterances).

\noindent \textbf{MuST-C En-De} MuST-C is a multilingual speech translation corpus extracted from the TED talks.
We run the experiments on the English-German speech translation dataset of 400 hours speech with 230K utterances.
We select the model on the dev set (1,408 utterances) and report results on the tst-COMMON set (2,641 utterances).

\noindent \textbf{Unrestricted Setting} We use the additional ASR and MT data for pre-training.
The 960 hours LibriSpeech ASR corpus is used for the English ASR model.
We extract 10M sentences pairs from the WMT14 English-French and 18M sentence pairs from the Opensubtitle2018\footnote{http://opus.nlpl.eu/OpenSubtitles-v2018.php} English-German translation datasets.

\noindent \textbf{Preprocessing} Followed the preprocessing recipes of ESPnet \cite{Inaguma_ACL2020}, we remove the utterances of more than 3,000 frames and augment speech data by speed perturbation with factors of 0.9, 1.0, and 1.1.
The 80-channel log-mel filterbank coefficients with 3-dimensional pitch features are extracted for speech data.
We use the lower-cased transcriptions without punctuations.
The text is tokenized using the scripts of Moses \cite{Koehn_ACL2007}.
We learn Byte-Pair Encoding \cite{Sennrich_acl2016} subword segmentation with 10,000 merge operations based on a shared source and target vocabulary for all datasets.

\subsection{Model Settings}

All experiments are implemented based on the ESPnet toolkit\footnote{https://github.com/espnet/espnet}.
We use the Adam optimizer with $\beta_1 = 0.9$, $\beta_2 = 0.997$ and adopt the default learning schedule in ESPnet.
We apply dropout with a rate of 0.1 and label smoothing $\epsilon_{ls} = 0.1$ for regularization.

For reducing the computational cost, the input speech features are processed by two convolutional layers, which have a stride of $2 \times 2$ and downsample the sequence by a factor of 4 \cite{Weiss_ISCA2017}.
The encoder consists of 12 layers for both the ASR and vanilla ST models, and 6 layers for the MT model.
The encoder of SATE includes an acoustic encoder of 12 layers and a textual encoder of 6 layers.
The decoder consists of 6 layers for all models.
The weight of CTC objective $\alpha$ for multitask learning is set to 0.3 for all ASR and ST models.
The coefficients $\beta$ and $\gamma$ are set to 0.5 in Eq. (\ref{eq:final_obj}) for the MTKD method.

Under the restricted setting, we employ the Transformer architecture, where each layer comprises 256 hidden units, 4 attention heads, and 2048 feed-forward size.
For the unrestricted setting, we use the superior architecture Conformer \cite{Gulati_ISCA2020} on the ASR and ST tasks and widen the model by increasing the hidden size to 512 and attention heads to 8.
The ASR\footnote{We use the pre-trained ASR model offered by ESPnet.} and MT models pre-train with the additional data and fine-tune the model parameters with the task-specific data.

During inference, we average the model parameters on the best 5 checkpoints based on the performance of the development set.
We use beam search with a beam size of 4 for all models.
Different from previous work, we report the case-sensitive SacreBLEU\footnote{BLEU+case.mixed+numrefs.1+smooth.exp+tok.13a\\+version.1.4.14} \cite{Post_wmt18} for future standardization comparison across papers.


\subsection{Results}

\begin{table}[t]
  \centering
  \begin{tabular}{l|c|c}
      \toprule
      Method & Restricted & Unrestricted \\
      \midrule
      ESPnet MT$^*$ & 27.63 & - \\
      ESPnet Cascaded$^*$ & 23.65 & - \\
      MT & 26.9 & 31.1 \\
      Cascaded ST & 23.3 & 28.1 \\
      \midrule
      ESPnet E2E ST$^*$ & 22.33 & - \\
      E2E ST & 22.1 & 23.6 \\
      \quad +Pre-training & 23.1 & 25.6 \\
      \midrule
      SATE & 23.3 & 23.6 \\
      \quad +Pre-training & 24.1 & 27.3 \\
      \quad\quad+MTKD & 24.7 & 27.9  \\
      \quad\quad\quad+SpecAug & \textbf{25.2} & \textbf{28.1} \\
      \bottomrule
  \end{tabular}
  \caption{BLEU scores [$\%$] on the test set of MuST-C En-De corpus. $*$: results reported in the ESPnet toolkit.}
  \label{must_c}
\end{table}

\noindent \textbf{Results on MuST-C En-De} Table \ref{must_c} summaries the experimental results on the MuST-C En-De task.
Under the restricted setting, the cascaded ST model translates the output of the ASR model, which degrades the performance compared with the MT model that translates from the reference transcription.
The performance of the E2E ST baseline with pre-training is only slightly lower than the cascaded counterpart.
SATE outperforms the baseline model significantly.
This demonstrates the superiority of stacked acoustic and textual encoding for the speech translation task.
Incorporating the pre-trained ASR and MT models into SATE releases the encoding burden of the model and achieves a remarkable improvement.
The MTKD method provides a strong supervised signal and forces the model to preserve the pre-trained knowledge.
Furthermore, we utilize the SpecAugment \cite{Park_ISCA2019} which is applied in the input speech features for better generalization and robustness\footnote{It is a fair comparison because the ASR model in the cascaded ST system also trains with the SpecAugment.}.
It yields a remarkable improvement of 1.9 BLEU points over the cascaded baseline and achieves a new state-of-the-art performance.

Under the unrestricted setting, the large-scale ASR and MT data is available, whereas the ST data is scarce.
This leads to the cascaded method outperforms the vanilla E2E method with a huge margin of 4.5 BLEU points.
The pre-training only slightly closes the gap due to the modeling deficiency and representation inconsistency.
SATE incorporates the pre-trained models fully, which achieves a significant improvement of 3.7 BLEU points.
With the MTKD and SpecAugment methods, we achieve a comparable performance of 28.1 BLEU points.
To our knowledge, we are the first to develop an end-to-end ST system that achieves comparable performance with the cascaded counterpart when large-scale ASR and MT data is available.

\begin{table}[t]
  \centering
  \begin{tabular}{l|c|c}
      \toprule
      Method & Restricted & Unrestricted \\
      \midrule
      ESPnet MT$^*$ & 18.09 & - \\
      ESPnet Cascaded$^*$ & 16.96 & - \\
      MT & 17.5 & 21.3 \\
      Cascaded ST & 16.3 & 20.6 \\
      \midrule
      ESPnet E2E ST$^*$ & 16.22 & - \\
      E2E ST & 16.7 & 17.7 \\
      \quad+Pre-training & 17.1 & 20.0 \\
      \midrule
      SATE & 17.6 & 18.1 \\
      \quad+Pre-training & 17.4 & \textbf{20.8} \\
      \quad\quad+MTKD & 17.7 & \textbf{20.8}  \\
      \quad\quad\quad+SpecAug & \textbf{18.3} & \textbf{20.8} \\
      \bottomrule
  \end{tabular}
  \caption{BLEU scores [$\%$] on the test set of LibriSpeech En-Fr corpus. $*$: results reported in the ESPnet toolkit.}
  \label{libri}
\end{table}

\noindent \textbf{Results on LibriSpeech En-Fr} Table \ref{libri} summaries the experimental results on the LibriSpeech En-Fr task.
Different from the MuST-C corpus, it is of small magnitude with clean speech data.
This results in that the performance of the vanilla E2E baseline is even better than the cascaded counterpart under the restricted setting.
Furthermore, pre-training helps the model achieve an improvement of 0.8 BLEU points over the cascaded baseline.
More interestingly, SATE without pre-training outperforms the above methods significantly, even achieves a slight improvement than the MT model.
A possible reason is that the diverse acoustic representation is fed to the textual encoder, which improves the robustness of the model.
This demonstrates the superiority of our method.

Combining our proposed methods yields a substantial improvement of 2.0 BLEU points over the cascaded baseline.
It is a new state-of-the-art result of 18.3 BLEU points.
Also, we outperform the cascaded counterpart by 0.2 BLEU points on the unrestricted task.

\section{Analysis}

\subsection{Model Performance vs. Speedup}

\begin{table}[t]
  \centering
  \begin{tabular}{l|c|c}
      \toprule
      Method & BLEU & RTF/Speedup \\
      \midrule
      Cascaded ST & 23.3 & 0.0286/$1.00\times$ \\
      \midrule
      E2E ST & 22.1 & \multirow{2}{*}{0.0150/$1.91\times$}  \\
      \quad+Pre-training & 23.1 &  \\
      \midrule
      E2E ST (Enc 18) & 22.8 & \multirow{2}{*}{0.0155/$1.85\times$} \\
      \quad+Pre-training & 23.5 &  \\
      \midrule
      SATE & 23.3 & \multirow{3}{*}{0.0169/$1.69\times$} \\
      \quad+Pre-training & 24.1 &  \\
      \quad+All & 25.2 &  \\
      \bottomrule
  \end{tabular}
  \caption{BLEU scores [\%] and speedup on the test set (2641 utterances) of the MuST-C En-De corpus under the restricted setting. We evaluate the RTF on the NVIDIA V100 GPU with a batch size of 4 for all models.}
  \label{bleu_speedup}
\end{table}

In Table \ref{bleu_speedup}, we summarize the performance and inference speedup based on the real time factor (RTF).
The vanilla E2E ST model yields an inference speedup of $1.91\times$ than the cascaded counterpart and demonstrates the low latency of the end-to-end methods.
We increase the encoder layers for comparison with SATE under the similar model parameters.
However, there is a remarkable gap of 0.5 or 0.6 BLEU points, with or without pre-training.

Our method not only improves the performance of 1.9 BLEU points but also reaches up to $1.69\times$ speedup than the cascaded baseline.
This encourages the application of the end-to-end ST model in practical scenarios.

\subsection{Effects of Pre-trained Modules}

\begin{table}[t]
  \centering
  \begin{tabular}{l|c|c}
      \toprule
      Pre-trained Module & MuST-C & LibriSpeech \\
      \midrule
      All & 27.3 & 20.8 \\
      \quad-ASR Enc & 24.7 & 19.9 \\
      \quad-MT & 25.1 & 19.4 \\
      \quad-MT Enc & 25.7 & 20.7 \\
      \quad-MT Dec & 25.3 & 19.9 \\
      \bottomrule
  \end{tabular}
  \caption{Effects of the pre-trained modules on BLEU scores [$\%$] under the unrestricted setting. We only remove one pre-trained module in each experiment.}
  \label{effect_modules}
\end{table}

\begin{table}[t]
  \centering
  \begin{tabular}{l|c|c}
      \toprule
      Design & MuST-C & LibriSpeech \\
      \midrule
      None & 25.7 & 21.7 \\
      Soft & 25.7 & 21.9 \\
      Mapping & 26.0 & 21.8 \\
      Fusion & 26.4 & 21.9 \\
      \bottomrule
  \end{tabular}
  \caption{BLEU scores [$\%$] of different adaptor setups on the development set under the unrestricted setting.}
  \label{effect_adaptor}
\end{table}

The effects of the pre-trained modules are shown in Table \ref{effect_modules}.
The model performance drops significantly without the pre-trained ASR encoder, especially on the MuST-C corpus that contains noisy speech.
The model parameters of pre-trained MT model are updated for adapting the output representation of the random initialized acoustic encoder.
This results in the catastrophic forgetting problem \cite{goodfellow_Corr2015}.
The effect of the pre-trained MT model is more remarkable on the LibriSpeech corpus due to the modeling burden on the translation.
The benefit of the pre-trained MT decoder is larger than the MT encoder.
This is contrary to the previous conclusions that the MT encoder helps the performance significantly \cite{Li_Corr2020}.
A possible reason is that the pre-trained ASR encoder provides a rich representation and acts as part of the MT encoder, this leads to lower performance degradation when the textual encoder trains from scratch.

Each pre-trained module has a great effect on the final performance.
With the complete integration of the pre-trained modules, the model parameters are updated slightly, which preserves the pre-trained knowledge.

\subsection{Effects of The Adaptor}

We show the effects of the adaptor in Table \ref{effect_adaptor}.
The straight connection which omits the representation inconsistency issue results in the lower benefit of pre-training.
Although the soft representation aims at generating the adaptive representation, there is no obvious improvement on the MuST-C corpus.
A possible reason is that the noisy speech inputs produce the misalignment probabilities, which disturbs the textual encoding.
The mapping method achieves a slight improvement by transforming the acoustic representation to the textual representation.
Fusing the soft and mapping representation enriches the information and avoids the representation inconsistency issue, which achieves the best performances.

\subsection{Impact on Localness}

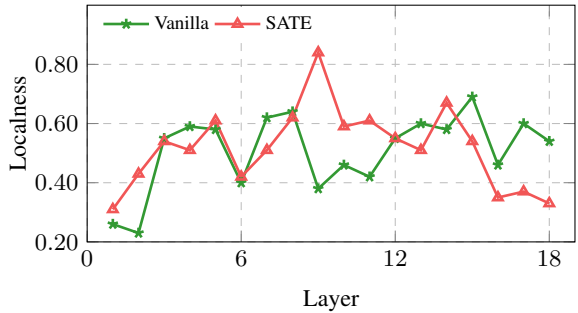
\begin{figure}[t!]
  \centering
    \begin{tikzpicture}
      \footnotesize{
      \begin{axis}[
        ymajorgrids,
        xmajorgrids,
        grid style=dashed,
        width=.5\textwidth,
        height=0.295\textwidth,
        legend columns=2,
        legend entries={Vanilla, SATE},
        legend style={
          draw=none,
          line width=1pt,
        },
        legend style={at={(0.25,1.0)}, anchor=north,
        nodes={scale=0.8, transform shape}},
        xmin=0, xmax=19,
        ymin=0.2, ymax=1,
        xtick={0, 6, 12, 18},
        ytick={0.2, 0.4, 0.6, 0.8},
        xlabel=\footnotesize{Layer},
        ylabel=\footnotesize{Localness},
        ylabel style={yshift=-1em},
        xlabel style={yshift=0.0em},
        yticklabel style={/pgf/number format/precision=2,/pgf/number format/fixed zerofill},
        scaled ticks=false,
        ]
    \addplot[ugreen!80, mark=star, line width=1pt] file {data/base_pre.txt};
    \addplot[myred!80, mark=triangle, line width=1pt] file {data/sate_pre.txt};
    \end{axis}
    }
  \end{tikzpicture}
  \caption{The localness of the vanilla E2E ST model and SATE model with pre-training.}
  \label{comparison}
\end{figure}

We show the encoder localness of the vanilla E2E ST model and SATE model with pre-training in Figure \ref{comparison}.
As mentioned above, the vanilla ST model inherits the preference of ASR, which focuses on short-distance dependencies.
SATE initializes with the pre-trained ASR and MT encoders, which stacks acoustic and textual encoding.
The complementary behaviors of the pre-trained models benefit the translation, that is, the lower layers act like an ASR encoder while the upper layers capture global representation like an MT encoder.

\section{Conclusion}

In this paper, we investigate the difficulty of speech translation and shed light on the reasons why pre-training has been challenging in ST.
This inspires us to propose a \emph{Stacked Acoustic-and-Textual Encoding} method,
which is straightforward to incorporate the pre-trained models into ST.
We also introduce an adaptor module and a multi-teacher knowledge distillation method for bridging the gap between pre-training and fine-tuning.

Results on the LibriSpeech and MuST-C corpora demonstrate the superiority of our method.
Furthermore, we achieve comparable or even better performance than the cascaded counterpart when large-scale ASR and MT data is available.

\section{Acknowledgement}

This work was supported in part by the National Science Foundation of China (Nos. 61876035 and 61732005), the National Key R\&D Program of China (No. 2019QY1801), and the Ministry of Science and Technology of the PRC (Nos. 2019YFF0303002 and 2020AAA0107900). The authors would like to thank anonymous reviewers for their comments.

\bibliographystyle{acl_natbib}
\bibliography{acl2021}

\end{document}